\documentclass{article}

\usepackage{arxiv}
\usepackage{float}
\usepackage{booktabs}
\usepackage[table,xcdraw]{xcolor}
\usepackage[utf8]{inputenc} 
\usepackage[T1]{fontenc}    
\usepackage{hyperref}       
\usepackage{url}            
\usepackage{booktabs}       
\usepackage{amsfonts}       
\usepackage{nicefrac}       
\usepackage{microtype}      
\usepackage{lipsum}
\usepackage{graphicx}
\graphicspath{ {./images/} }

\title{Improved prediction rule ensembling through model-based data generation}

\author{
 Benny Markovitch \\
  EBMT (European Society for Blood and Marrow Transplantation), Leiden\\
  \texttt{b.markovitch.correspond@gmail.com} \\
   \And
   Marjolein Fokkema\\
  Leiden University \\
  \texttt{m.fokkema@fsw.leidenuniv.nl} \\
}

\begin{document}
\maketitle
\begin{abstract}
Prediction rule ensembles (PRE) provide interpretable prediction models with relatively high accuracy. PRE obtain a large set of decision rules from a (boosted) decision tree ensemble, and  achieves sparsity through application of Lasso-penalized regression. This article examines the use of surrogate models to improve performance of PRE, wherein the Lasso regression is trained with the help of a massive dataset generated by the (boosted) decision tree ensemble. This use of model-based data generation may improve the stability and consistency of the Lasso step, thus leading to improved overall performance. We propose two surrogacy approaches, and evaluate them on simulated and existing datasets, in terms of sparsity and predictive accuracy. The results indicate that the use of surrogacy models can substantially improve the sparsity of PRE, while retaining predictive accuracy, especially through the use of a nested surrogacy approach. 
\end{abstract}


\section{Introduction} The tradeoff between a model’s interpretability and its prediction accuracy is a major issue in the field of statistical learning \cite{YarkyWest17}. When facing this tradeoff, modellers aiming to obtain an interpretable prediction model typically have two options \cite{GuidyMonr18}: The first option involves the fitting of a highly accurate black-box model, which is then interpreted using a so-called surrogate model, which mimics the black-box model using a simpler, easier to interpret model  such as a decision tree \cite{CravyShav95}. The second option involves the use of a more interpretable model to begin with, such as logistic regression. Unfortunately, these approaches do not fully resolve the accuracy-interpretability tradeoff. With the surrogacy approach, higher fidelity to the black box model can come at the cost of higher complexity, and thus lowered interpretability \cite{MingQu18}. Using a  more interpretable  model, on the other hand, generally  leads to lower accuracy than what is typically obtained using black box models \cite{YarkyWest17}.

Prediction Rule Ensemble (PRE) algorithms like RuleFit \cite{FrieyPope08}, Node harvest \cite{Mein10}, and C4.5 \cite{Quin14} provide promising interpretable statistical learning methods, in which an initial rule ensemble is obtained from the nodes of a decision tree ensemble, and from which a smaller subset of prediction rules are selected. In the current paper, we will focus on the RuleFit algorithm, which employs tree boosting to obtain an initial ensemble, whose prediction rules are then selected and weighed using the Lasso \cite{Tibs96}. More specifically, we focus on the R implementation of \cite{Fokk20}. The RuleFit procedure tends to result in slightly lower prediction accuracy than that of methods such as random forests and boosting, while the resulting model is considerably easier to interpret \cite{FrieyPope08, ShimyLi14, Fokk20}.

Although PRE use a simplification of a black-box model (boosting), it is typically approached as an interpretable statistical learning model, rather than as a surrogate model. Yet since it shares important attributes with surrogate models, it might benefit from a consideration of the surrogacy approach. A common surrogacy approach is the pedagogical approach \cite{GuidyMonr18, MingQu18}, in which a large dataset is generated based on the distributions of the training dataset’s features, combined with the predictions given by the black box model (known as the Oracle). The resulting dataset is then typically used to train a decision tree whose inherent instability is mitigated by the large volume of generated data. Similarly, when building PRE, the Boosting model could be used as an Oracle that generates a large volume of data, which might improve the stability and consistency of the Lasso rule-selection step.

\subsection{Surrogacy within PRE}
The high interpretability of PRE is due to its sparsity, which is obtained by using the Lasso to select from the rules generated by the Boosting model. However, the Lasso comes with disadvantages that could harm the performance of PRE. According to the ‘No-free-lunch Theorem’ of Xu, Caramanis, and Mannor \cite{XuyCara11}, algorithms that identify and remove redundant features are more sensitive to random variations in the training data, making them unstable. Xu, Caramanis, and Mannor have shown that the Lasso indeed suffers from this trade-off between sparsity and stability, such that its sparsity comes at the cost of instability. The Lasso’s cross-validation (CV) procedure is an additional source of algorithmic instability, due to variability inherent to its resampling procedure \cite{LimyYu16}. Furthermore, Ali and Tibshirani (\cite{AliyTibs19}; see also \cite{Tibs96}) have shown that if the number of features ($P$) is bigger than the size of the training sample ($N$), and not all features are continuous, then the Lasso may not have a unique optimal solution, making the Lasso inconsistent. The Lasso’s instability and inconsistency are expected to decrease the performance of PRE, unless the training dataset is large enough to mitigate the Lasso’s inherent instability, and to solve the Lasso’s inconsistency by having $N$ that is larger than the number of Boosting rules (which often ranges in the thousands). In other words, the Lasso’s disadvantages will more strongly harm PRE when small datasets are used for model fitting.

Taking a surrogacy-like data-generation approach within PRE, in which the Boosting model is used as a prediction-generating Oracle, could help mitigate the Lasso’s disadvantages when using small datasets. First, in cases where $N < P$ (where $P$ is the number of Boosting rules), the Lasso’s inconsistency should be resolved by a sufficiently large generated dataset of size $N_{Gen}$. Second, by generating a dataset with large $N_{Gen}$, the instability attributable to the CV’s resampling should be reduced. Third, a large $N_{Gen}$ should also mitigate the instability attributable to the Lasso’s sparsity. In addition, since the relationship between the generated dataset’s features and the Boosting model’s predictions is noiseless and deterministic, the Lasso’s instability could be further reduced. As such, the implementation of the proposed surrogate Lasso within PRE should result in more stable and consistent models, whose predictions closely mimic the Boosting model’s predictions.

While the use of the surrogate Lasso is expected to mitigate the regular Lasso’s inconsistency and instability, this will likely come at a cost. The typical data generation method (see implementation details in Section 5.1) leads to the loss of the features’ joint distribution, which results in an artificial decorrelation of the rules. While this decorrelation of rules could further stabilise the surrogate Lasso \cite{Fu98, DormyElit13}, it may also come at the cost of increased complexity, as highly-correlated rules could be more likely to be maintained than otherwise. Furthermore, the mere increase in $N_{Gen}$ could result in more rules being kept by the surrogate Lasso, even if their added predictive value is minimal \cite{Tibs96}. Lastly, the Boosting model is not designed for variable selection \cite{Frie02}, and thus its predictions are expected to be influenced by noise variables. This implies that a surrogate Lasso, which mimics the Boosting model, may be less likely to exclude noise variables than the regular Lasso. As such, while the surrogate Lasso is expected to be more consistent and stable than the regular Lasso, this is likely to come at the cost of greater model complexity, and more frequent inclusion of noise variables.

Although the surrogate Lasso may be more likely to include  noise variables and unnecessary rules than the regular Lasso, its rule-selection step should still possess desirable properties, potentially making it useful for a nested rule-selection procedure. This is because the surrogate Lasso’s large $N_{Gen}$ should make it well-suited for detecting useful rules, assuming that the Boosting model makes high-quality predictions. Yet, as mentioned before, the surrogate Lasso is expected to be sensitive to the Boosting model’s reliance on noise variables. Thus, the surrogate Lasso should show high sensitivity to useful rules, while lacking specificity when it comes to redundant rules. However, its specificity might still be high enough to filter out many of the redundant Boosting rules, while its high sensitivity should allow it to keep most useful rules. Thus, the surrogate Lasso might select a set of rules which is significantly smaller than the number of Boosting rules, at a minimal cost to the inclusion of potentially useful rules. Making it a potentially useful first selection step in a nested, two-step selection procedure.

In the proposed nested procedure, a Lasso model will use the original training dataset to select from the rules kept by the surrogate Lasso. In comparison to the surrogate Lasso, the nested Lasso should better account for the features’ joint distribution, be less sensitive to the Boosting model’s reliance on noise variables, and its sparsity should be less dependent on $N_{Gen}$. Furthermore, if the surrogate Lasso is allowed to pick no more rules than the original dataset’s $N$, then the nested Lasso will be consistent in the sense that it should reach its optimal solution given the data. In addition, by selecting from fewer rules than the regular Lasso, the instability brought by the Lasso’s sparsity and cross-validation resampling might be partially mitigated, as algorithmic instability is worsened in increasingly high-dimensional settings \cite{LimyYu16}. Thus, the nested Lasso might be more stable and consistent than the regular Lasso, without the surrogate Lasso’s increased complexity and reliance on noise variables.

\subsection{Current study}
This study’s main research question is whether the inner surrogacy approach can improve the performance of PRE. More specifically, since the Lasso’s inconsistency and instability are assumed to harm the regular Lasso’s performance, we hypothesize that the two surrogacy-based Lasso approaches yield improved performance compared to the regular approach. Ideally, this would translate into improved predictive accuracy. Furthermore, greater stability should result in improved rule selection \cite{LimyYu16}, such that the rules selected by the nested Lasso might be of higher quality and lower quantity than those selected by the regular Lasso. In what follows, we will first describe the proposed surrogacy implementations for PRE, after which we empirically evaluate and compare their predictive accuracy, rule quality and sparsity.

\section{Proposed algorithms}

\subsection{Data generation for surrogacy models}

This algorithm is inspired by Craven and Shavlik’s \cite{CravyShav95} data generation procedure: Given a set of features, data generation will be performed by resampling each feature, independently of all other features, into a pre-specified number of rows, and then binding all the resulting columns together. This maintains each feature’s marginal distribution, yet loses the features’ joint distribution. Following that, only unique rows will be maintained, in order to improve the ratio between the dataset’s information and the resulting $N_{Gen}$. Lastly, a pre-specified Oracle will be used to generate  a prediction per row, resulting in a generated dataset that is suitable for model fitting.

\subsection{Rule generation}Based on the PRE algorithm \cite{FrieyPope08, Fokk20}, a Boosting model will be fitted on a training dataset. The building blocks of the Boosting model are decision trees, which represent a recursive binary decision-making process \cite{BuhlyHoth07}. PRE’s default Boosting model additively builds 500 Conditional Inference Trees (ctrees; \cite{HothyHorn06}) with maximal depth of 3, and $\alpha = .05$, using the partykit’s package ctree function \cite{HothyZeil15}.

The Boosting algorithm additively builds a stable model from simple base learners which are highly biased alone, yet can be combined to make predictions with low bias \cite{BuhlyHoth07}. Ctrees with maximal depth of 3 fit this purpose, and make the resulting rules simple enough to be interpretable as parts of an ensemble. The algorithm implemented employs a gradient Boosting approach, in which ctrees are trained on pseudo response $\tilde{y}_b$, which is based on outcome variable $y$ and the base learners trained in previous iterations. With a continuous outcome, the pseudo response in iteration $b$ ($b = 1, \dots, 500$) is given by:  

\begin{eqnarray}
\tilde{y}_b = y - \sum^{b-1}_{m=1} \nu \times f_m(X)
\end{eqnarray}

where $\nu > 0$ is the user-specified learning rate which determines each tree’s influence  ($\nu = 0.01$ by default), and $f_m(X)$ is the prediction made by tree $m$ ($m = 1, \dots, b-1$) given feature set $X$. This amounts to the gradient boosting algorithm of Friedman (\cite{Frie01}; Algorithm 1), but omitting the line search (cf. \cite{BuhlyHoth07}; section 2.2). For binary classification, a Newton (instead of gradient) boosting approach is employed \cite{Sigr21, Fokk20}.

PRE employs a stochastic Boosting approach \cite{Frie02} in which each  tree is grown on a subsample of 50\% of the training data, which tends to lower the number of Boosting rules and reduce the prevalence of redundant rules \cite{DeBiyJani16}. Yet despite the higher prevalence of redundant rules, our initial experiments suggested that bootstrapping improves the Boosting model’s predictive accuracy on test observations. Since the Boosting model is used as an Oracle, making its prediction accuracy particularly important, it will always be trained using the bootstrap.

\subsection{Rule selection}
Once the Boosting algorithm is complete, all the fitted trees’ decision nodes will be used to form an ensemble of rules. From this ensemble, only unique rules and rules that are not collinear with previous rules will be kept. Furthermore, linear terms will be added in order to ease the approximation of linear relationships, and continuous features will be winsorized at the .025 and .975 quantiles \cite{FrieyPope08, Fokk20}. When used as linear terms, features will be divided by their standard deviation, and then multiplied by 0.4, in order to avoid being penalized by the Lasso for having lower variance than most rules \cite{FrieyPope08, Fokk20}. No further standardization or normalization will be needed for the Lasso. The Lasso will take the Boosting model’s rules as a set of dummy variables, alongside the aforementioned linear terms. It will then use penalized regression to select rules and linear terms from the feature set. This will be done using 5-fold cross-validation to find an optimal shrinkage parameter $\lambda $, which controls the number of terms maintained, and the magnitude of their associated slopes. The actual $\lambda$, picked to create the final model will be the highest $\lambda$, value whose loss is no more than 1 standard error away from the minimal loss. The regression terms maintained by the Lasso will then be used as a sparse and interpretable prediction model. Note that 5-fold CV will be used, instead of the default 10, to reduce computational demands.

We will examine three variations on the Lasso algorithm:

\begin{itemize}
  \item \relax The regular Lasso, which will take the Boosting model's rules and the linear terms as features set, and then use the training data to select a sparser model.
  \item \relax The surrogate Lasso, which will take the same starting feature set as the regular Lasso, while using a generated dataset in order to select a sparser model (using 3-fold CV to reduce computational burden). This will be extended into a 2nd-level surrogate Lasso, which will use a new dataset generated in the same manner as before to select from the features kept by the 1st-level surrogate Lasso. Our initial work showed that this 2nd-level surrogate Lasso leads to reduced model complexity, at no cost to accuracy. 
  \item \relax The nested Lasso will use the same training data as the regular Lasso, yet will use a 5-fold CV only to select from the rules kept by the 1st-level surrogate Lasso. Initial experiments suggested that 3-step selection, as opposed to 2, only adds computational load, at no benefit to accuracy or sparsity.
\end{itemize}
  
\section{Empirical evaluation}

\subsection{Datasets}The  experiments will rely on 5 datasets, 3 of which have been previously used for evaluating  PRE \cite{Fokk20}. These datasets will only include regression and binary classification problems. All datasets, except the life expectancy dataset, come from R’s mlbench package \cite{LeisyDimi21}, and are described in more detail below.

\paragraph{Wisconsin Breast Cancer dataset:} This dataset will be used for binary classification, and also for regression with a simulated outcome variable. In the original dataset, 9 continuous features are used to predict whether a patient's cancer is benign or malignant. These same features will also be used as predictors when simulating a continuous outcome which is a linear function of several features (See experimental details in Section 4.1). The dataset includes 683 observations after exclusion of rows with missing values.

\paragraph{Johns Hopkins University Ionosphere dataset:} This dataset will be used for binary classification. It contains one binary feature, and 32 continuous features, which are used to predict the type of high-energy structures in the atmosphere, which have been simplified into two categories: good or bad. The dataset contains 351 observations.

\paragraph{Metal vs. Rock dataset:} This dataset will be used for binary classification. It contains 60 continuous features which are used to predict whether a material is rock or metal. This dataset will also be used for regression, where 59 of its continuous features will be used to predict the values of a randomly-selected feature. The dataset includes 208 observations.

\paragraph{Boston Housing dataset:} This dataset will be used for regression. It contains one binary feature, and 12 continuous features, which are used to predict a house's price. The dataset includes 506 observations.

\paragraph{Life expectancy dataset:} This dataset was obtained via Kaggle, and contains 1 categorical predictor, and 19 continuous predictors, which are used to predict a country's life expectancy at a given year. Note that the country was excluded as a predictor, in order to ensure the model can predict life expectancy in countries that did not appear in the training set. After omitting rows with empty values, the dataset included 1649 observations.

\subsection{Performance indicators}For each experiment's results, confidence intervals will be based on the distribution of paired-samples differences, where mean differences ($\bar{D}$) are assessed as 

\begin{eqnarray}
\frac{1}{N} \sum_{i=1}^N (y_{i1} - y_{i2})
\end{eqnarray}

where $i$ reflects the experimental iteration, $N$ reflects the number of iterations, and $y_{i1}$ reflects the value of interest of method 1 (and $y_{i2}$ of method 2). Thus, the variance of the differences distribution will be assessed as $\mathrm{var}(Y_1) + \mathrm{var}(Y_2) - 2 \mathrm{cov}(Y_1,Y_2)$, leading to the standard error of differences being $\sqrt{\frac{\mathrm{var}(\bar{D})}{N}}$.

Differences will be treated as statistically significant if their 95\% confidence intervals exclude 0, or, when confidence intervals are not available, if the associated two-tailed p-value is lower than .05. Iterations in which one of the Lassos creates an intercept-only model will be excluded to facilitate comparability.

\paragraph{Test set accuracy:} In the regression case, test accuracy will be assessed as mean squared difference (MSE) between predicted test set values, and true test values. In the classification case, test accuracy will be evaluated as the correct classification rate (CCR) given the predicted test set class labels, and the true test set class labels.

 \paragraph{Model complexity:} Will be measured as the number of terms maintained by each model, with more terms resulting in higher complexity.

\paragraph{Quality of parameters:} When the outcome is not simulated using a known function, the quality of selected terms will only be calculated as a function of model accuracy, and model complexity. In the regression case, this is measured as $\frac{\mathrm{ESS}}{T} $, where ESS refers to the model's explained sum of squares on the test outcome and $T$ refers to the number of selected terms. As such, this quantity reflects how ESS increases, on average, per term. In the binary classification case, the quality of selected terms is measured as $\frac{\mathrm{CCR}}{T}$, which reflects the correct classification rate increment, on average, per term.

When the outcome is simulated from a known function, the quality of selected terms will also be estimated as the correlation between features’ standardized model importances (see \cite{Fokk20, FrieyPope08}) and the slopes assigned to them as part of the simulation. Furthermore, true and false positive rates will be calculated to assess the quality of variable selection. True positive rates will be estimated as the intersection between the variables utilized by PRE and the variables used to generate the outcome of the simulation algorithm. False positive rates will be estimated as the intersection between the variables utilized by PRE and the variables not used to generate the outcome of the simulation algorithm.

 \paragraph{Model stability:} will be assessed in terms of the stability of feature selection, and will be estimated using Nogueira's Index \cite{NogeySechi17} which ranges from -1 to 1, with higher values suggestive of higher stability of feature selection. These values will be reported in the Appendix.

  \paragraph{Computational load:} total computation time in seconds will be calculated for each procedure, and will be reported in the Appendix. Note that since the experiments will be performed on different computers, computation times should be compared within experiments, but not between experiments.

\section{Experiments}
Unless otherwise specified, the upcoming experiments share the following characteristics: 1) whenever data generation is used, $N_{Gen} = 10,000$, 2) only real data are used, with the dataset's originally intended outcome variable, 3) PRE are run with regular Lasso, surrogate Lasso, and nested Lasso, and 4) the experiment is iterated 200 times, each iteration involving random division of the dataset into a training dataset and a testing dataset of equal size. The results of all iterations will then be analyzed together.

\subsection{Experiment 1: simulated linear relationship using the Breast Cancer dataset}In this experiment, outcome variable Y will be simulated using the dataset's 9 continuous features. In each simulation, 3 features will be randomly selected to relate to Y in accordance to

\begin{eqnarray}
Y = \sum^{3}_{k=1} B_k \times X_k + \epsilon
\end{eqnarray}
 
where the influence ($B_k$) of each of the selected features ($X_k$) on the outcome variable will be determined by a uniform distribution $\mathcal{U}(0.5, 2.0)$, multiplied by a random variable which takes values \{-1, 1\} according to a binomial distribution in which each sign is equally likely to be picked. In addition, noise ($\epsilon$) will be independently generated from a normal distribution $\mathcal{N}(0, 25)$.

After generating $Y$, the dataset's observations will be split into halves, a test set, and a training set on which the Boosting model will be fit. The Lasso models will then utilize the Boosting model's decision rules in their rule-selection step(s), but will not use linear terms for each feature, as PRE are known to struggle when modelling linear relations without linear terms \cite{FrieyPope08}. As such, the linear relationship is a modeling challenge for PRE when only prediction rules are used.

Since the function responsible for the outcome is known, quality of terms will also be assessed in terms of correlation between features' standardized model importances and the slopes assigned to them as part of the simulation. As a secondary measure of quality of terms, true and false positive rates of variable selection will also be assessed, instead of variable selection stability. Aside from these clarifications, the measurements will be the same as described above (Section 3.2).

All 200 iterations converged without issues. Means and standard deviations of accuracy MSE, number of terms, and average accuracy ESS increment per term are depicted in Table 2. The distributions of accuracy MSE and number of terms are shown, per method, in Figure 1. The regular Lasso had the highest accuracy MSE, which was statistically distinguishable from the nested Lasso (95\% CI [0.23-0.53]), which itself had higher accuracy MSE than the surrogate Lasso (95\% CI [1.28-1.68]). When it came to the number of terms, the regular and nested Lassos were statistically indistinguishable, and their ESS increment per variable was similarly indistinguishable.

\begin{table}[H]
\centering
\caption{{Summary of results for simulated linear relationships based on the Breast Cancer dataset. Values represent means over 200 iterations (standard deviations in parentheses). In each column, the result of the best performing method (excluding boosting) is printed in bold.} }
\begin{tabular}{@{}lcccl@{}}
\toprule
\cellcolor[HTML]{000000}{\color[HTML]{333333} \cellcolor[HTML]{FFFFFF}\textbf{Procedure}} &
  \cellcolor[HTML]{FFFFFF}\textbf{Accuracy MSE} &
  \cellcolor[HTML]{FFFFFF}\textbf{Terms} &
  \cellcolor[HTML]{FFFFFF}\textbf{Accuracy ESS per term} &
   \\ \midrule
\textbf{Boosting}        & 27.08 (2.08)          & 613 (167)             & NA              &  \\
\textbf{Regular Lasso}   & 28.98 (2.40)          & \textbf{22.63 (9.21)} & 514.91 (348.17) &  \\
\textbf{Surrogate Lasso} & \textbf{27.11 (2.07)} & 127.74 (26.99)        & 105.65 (90.20)  &  \\
\cellcolor[HTML]{FFFFFF}\textbf{Nested Lasso} &
  \cellcolor[HTML]{FFFFFF}28.60 (2.45) &
  \cellcolor[HTML]{FFFFFF}23.16 (7.92) &
  \cellcolor[HTML]{FFFFFF}\textbf{518.64 (378.66)} &
   \\ \bottomrule
\end{tabular} 
\end{table}

\begin{figure}[H] 
    \centering
    \includegraphics[width=\linewidth, height=7cm]{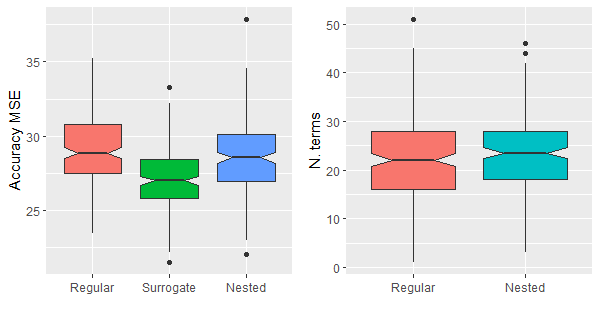}
        \caption{{Boxplots of the distributions of accuracy MSE, and number of terms, per Lasso type. Notches represent the interval covered by $1.58\times\frac{IQR}{\sqrt N}$, with IQR referring to the interquartile range.}}
\end{figure}

\subsection{Experiment 2: Boston Housing data (regression)}
All 200 iterations worked without issues. Results are presented in Table 3. The surrogate and regular Lasso did not significantly differ in accuracy, yet both had lower accuracy MSE than the nested Lasso. For example, the 95\% confidence interval of the differences between the regular and nested Lassos' was 0.146 to 0.694. Furthermore, the regular Lasso had significantly more terms than the nested Lasso (95\% CI [13-17.83]). As such, the nested Lasso had greater accuracy ESS increment per term than the regular Lasso (95\% CI [56.05-78.49]). The distributions of accuracy MSE and number of terms are shown, per method, in Figure 2.

\begin{table}[H]
\centering
\caption{{Summary of results based on the Boston Housing dataset. Values represent means over 200 iterations (standard deviations in parentheses). In each column, the result of the best performing method (excluding boosting) is printed in bold.} }
\begin{tabular}{@{}lcccl@{}}
\toprule
\cellcolor[HTML]{FFFFFF}{\color[HTML]{333333} \textbf{Procedure}} &
  \cellcolor[HTML]{FFFFFF}\textbf{Accuracy MSE} &
  \cellcolor[HTML]{FFFFFF}\textbf{Terms} &
  \cellcolor[HTML]{FFFFFF}\textbf{Accuracy ESS per term} &
   \\ \midrule
\textbf{Boosting}        & 14.39 (3.18)          & 1867 (97)              & NA             &  \\
\textbf{Regular Lasso}   & 15.06 (3.72)          & 67.02 (17.69) & 279.43 (79.47) &  \\
\textbf{Surrogate Lasso} & \textbf{14.96 (3.14)} & 235.87 (7.74)          & 74.21 (7.71)   &  \\
\cellcolor[HTML]{FFFFFF}\textbf{Nested Lasso} &
  \cellcolor[HTML]{FFFFFF}15.48 (3.69) &
  \cellcolor[HTML]{FFFFFF}\textbf{51.61 (8.93)} &
  \cellcolor[HTML]{FFFFFF}\textbf{346.7 (70.18)} &
   \\ \bottomrule
\end{tabular}
\end{table}

\begin{figure}[H] 
    \centering
    \includegraphics[width=\linewidth, height=7cm]{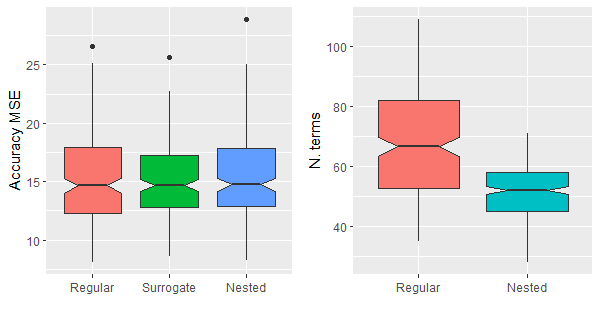}
        \caption{{Boxplots of the distributions of accuracy MSE, and number of terms, per Lasso type. Notches represent the interval covered by$1.58\times\frac{IQR}{\sqrt N} $, with IQR denoting interquartile range.}}
\end{figure}

\subsection{Experiment 3: Metal vs Rock dataset (regression)}
A total of 198 out of 200 iterations worked without issues; 2 iterations involved intercept-only models, and were thus excluded. Means and standard deviations of accuracy MSE, number of terms, and average accuracy ESS per term are depicted in Table 4. The nested and regular Lassos had statistically indistinguishable accuracy, while the surrogate Lasso had significantly lower accuracy MSE than both. For example, the 95\% CIs of the MSE differences between the nested and surrogate Lassos ranged from .0005 to .0007. The regular Lasso had significantly more terms than the nested Lasso (95\% CI [21.09-25.53]). As a result, the nested Lasso had greater accuracy ESS increment per term than the regular Lasso (95\% CI [.0386-.0523]). The distributions of accuracy MSE and number of terms are shown, per method, in Figure 3.

\begin{table}[H]
\centering
\caption{{Summary of results based on the Metal vs Rock dataset. Values represent means over 198 iterations (standard deviations in parentheses). In each column, the result of the best performing method (excluding boosting) is printed in bold.} }
\begin{tabular}{@{}lcccl@{}}
\toprule
\cellcolor[HTML]{FFFFFF}{\color[HTML]{333333} \textbf{Procedure}} &
  \cellcolor[HTML]{FFFFFF}\textbf{Accuracy MSE} &
  \cellcolor[HTML]{FFFFFF}\textbf{Terms} &
  \cellcolor[HTML]{FFFFFF}\textbf{Accuracy ESS per term} &
   \\ \midrule
\textbf{Boosting}        & .00452 (.0034)          & 1054 (268)             & NA           &  \\
\textbf{Regular Lasso}   & .00539 (.004)           & \textbf{49.76 (17.43)} & .0484 (.046) &  \\
\textbf{Surrogate Lasso} & \textbf{.00480 (.0036)} & 93.77 (6.93)           & .0294 (.027) &  \\
\cellcolor[HTML]{FFFFFF}\textbf{Nested Lasso} &
  \cellcolor[HTML]{FFFFFF}.00538 (.004) &
  \cellcolor[HTML]{FFFFFF}\textbf{26.45 (6.73)} &
  \cellcolor[HTML]{FFFFFF}\textbf{.0938 (.086)} &
   \\ \bottomrule
\end{tabular}
\end{table}

\begin{figure}[H] 
    \centering
    \includegraphics[width=\linewidth, height=7cm]{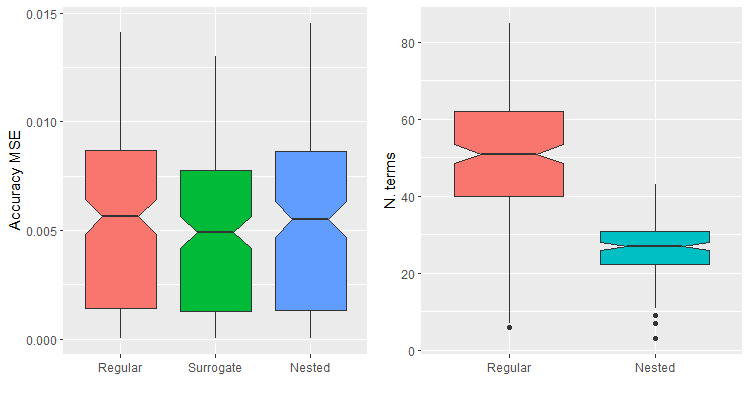}
        \caption{{Boxplots of the distributions of accuracy MSE, and number of terms, per Lasso type. Notches represent the interval covered by$1.58\times\frac{IQR}{\sqrt N} $, with IQR denoting interquartile range.}}
\end{figure}

\subsection{Experiment 4: Life Expectancy dataset (regression)}
All iterations worked without issues. Means and standard deviations of accuracy MSE, number of terms, and average accuracy ESS per term are depicted in Table 5. The regular Lasso had the lowest MSE, followed by the nested Lasso's MSE (95\% CI [0.30-0.24]), which itself was lower than the surrogate Lasso's MSE (95\% CI [1.11-1.21]). Whereas the regular Lasso had significantly more terms than the nested Lasso (95\% CI [18.56-21.83]). Consequently, the nested Lasso had greater ESS increment per term than the regular Lasso (95\% CI [93.21-109.33]). The distributions of accuracy MSE and number of terms are shown, per method, in Figure 4.

\begin{table}[H]
\centering
\caption{{Summary of results based on the Life Expectancy dataset. Values represent means over 200 iterations (standard deviations in parentheses). In each column, the result of the best performing method (excluding boosting) is printed in bold.}}
\begin{tabular}{@{}lcccl@{}}
\toprule
\cellcolor[HTML]{FFFFFF}{\color[HTML]{333333} \textbf{Procedure}} &
  \cellcolor[HTML]{FFFFFF}\textbf{Accuracy MSE} &
  \cellcolor[HTML]{FFFFFF}\textbf{Terms} &
  \cellcolor[HTML]{FFFFFF}\textbf{Accuracy ESS per term} &
   \\ \midrule
\textbf{Boosting}        & 6.32 (0.43)          & 3188 (93.15)            & NA            &  \\
\textbf{Regular Lasso}   & \textbf{5.33 (0.41)} & {118.31 (11.65)} & 507.57 (52.5) &  \\
\textbf{Surrogate Lasso} & {6.76 (0.47)} & 636.83 (36.33)          & 91.81 (5.70)  &  \\
\cellcolor[HTML]{FFFFFF}\textbf{Nested Lasso} &
  \cellcolor[HTML]{FFFFFF}5.60 (0.42) &
  \cellcolor[HTML]{FFFFFF}\textbf{98.10 (9.08)} &
  \cellcolor[HTML]{FFFFFF}\textbf{608.83 (56.21)} &
   \\ \bottomrule
\end{tabular}
\end{table}

\begin{figure}[H] 
    \centering
    \includegraphics[width=\linewidth, height=7cm]{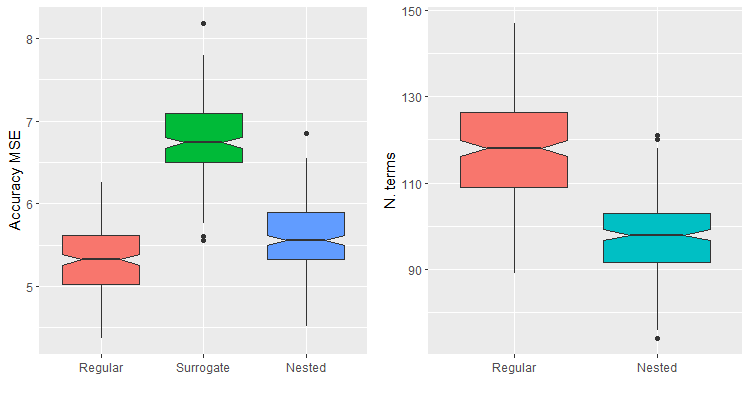}
        \caption{{Boxplots of the distributions of accuracy MSE, and number of terms, per Lasso type. Notches represent the interval covered by$1.58\times\frac{IQR}{\sqrt N} $, with IQR denoting interquartile range.}}
\end{figure}

\subsection{Experiment 5: Breast Cancer dataset (classification)}All 200 iterations worked without issues. Means and standard deviations of accuracy CCR, number of terms, and average accuracy CCR increment per term are depicted in Table 5. The regular and surrogate Lassos CCR's were statistically indistinguishable, while the nested Lasso had significantly higher accuracy CCR than both. For example, the 95\% confidence intervals of the differences between the nested and regular Lassos ranged from .0001 to .0014. No statistically significant differences were observed between the nested and regular Lassos' numbers of terms, nor in the accuracy CCR increment per term. The distributions of accuracy CCR and number of terms are shown, per method, in Figure 5.

\begin{table}[H]
\centering
\caption{{Summary of results based on the Breast Cancer dataset. Values represent means over 200 iterations (standard deviations in parentheses). In each column, the result of the best performing method (excluding boosting) is printed in bold.}}
\begin{tabular}{@{}lcccl@{}}
\toprule
\cellcolor[HTML]{FFFFFF}{\color[HTML]{333333} \textbf{Procedure}} &
  \cellcolor[HTML]{FFFFFF}\textbf{Accuracy CCR} &
  \cellcolor[HTML]{FFFFFF}\textbf{Terms} &
  \cellcolor[HTML]{FFFFFF}\textbf{Accuracy CCR per term} &
   \\ \midrule
\textbf{Boosting}        & .9638 (.0083)          & 1186 (136)            & NA          &  \\
\textbf{Regular Lasso}   & {.9633 (.0084)} & {15.57 (3.04)} & .064 (.013) &  \\
\textbf{Surrogate Lasso} & {.9631 (.0084)} & 162.41 (28.94)        & .006 (.001) &  \\
\cellcolor[HTML]{FFFFFF}\textbf{Nested Lasso} &
  \cellcolor[HTML]{FFFFFF}\textbf{.9640 (.0078)} &
  \cellcolor[HTML]{FFFFFF}\textbf{15.26 (2.91)} &
  \cellcolor[HTML]{FFFFFF}\textbf{.066 (.013)} &
   \\ \bottomrule
\end{tabular}
\end{table}

\begin{figure}[H] 
    \centering
    \includegraphics[width=\linewidth, height=7cm]{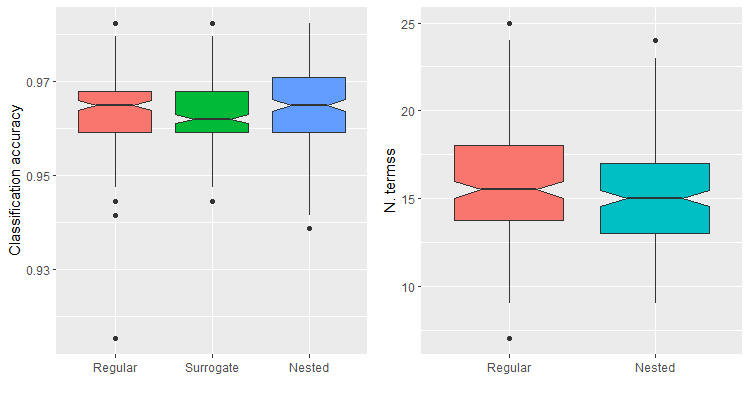}
        \caption{{Boxplots of the distributions of accuracy CCR, and number of terms, per Lasso type. Notches represent the interval covered by$1.58\times\frac{IQR}{\sqrt N} $, with IQR denoting interquartile range.}}
\end{figure}

\subsection{Experiment 6: Ionosphere dataset (classification)}
All 200 iterations worked without errors. Means and standard deviations of accuracy CCR, number of terms, and average accuracy CCR increment per term are depicted in Table 7. The only statistically significant difference in test accuracy was between the regular and surrogate Lassos, with the regular Lasso having slightly higher accuracy CCR (95\% CI [.002-.007]). The regular Lasso had significantly more terms than the nested Lasso (95\% CI [2.51-3.86]). Consequently, the nested Lasso had higher accuracy CCR increment per term (95\% CI [.006-.010]). The distributions of accuracy CCR and number of terms are shown, per method, in Figure 6.

\begin{table}[H]
\centering
\caption{{Summary of results based on the Ionosphere dataset. Values represent means over 200 iterations (standard deviations in parentheses). In each column, the result of the best performing method (excluding boosting) is printed in bold.
}}
\begin{tabular}{@{}lcccl@{}}
\toprule
\cellcolor[HTML]{FFFFFF}{\color[HTML]{333333} \textbf{Procedure}} &
  \cellcolor[HTML]{FFFFFF}\textbf{Accuracy CCR} &
  \cellcolor[HTML]{FFFFFF}\textbf{Terms} &
  \cellcolor[HTML]{FFFFFF}\textbf{Accuracy CCR per term} &
   \\ \midrule
\textbf{Boosting}        & .915 (.02)           & 1202.51 (131)         & NA           &  \\
\textbf{Regular Lasso}   & \textbf{.918 (.022)} & {19.62 (4.79)} & .05 (.058)   &  \\
\textbf{Surrogate Lasso} & {.914 (.021)} & 161.56 (6.18)         & .006 (.0002) &  \\
\cellcolor[HTML]{FFFFFF}\textbf{Nested Lasso} &
  \cellcolor[HTML]{FFFFFF}{.917 (.022)} &
  \cellcolor[HTML]{FFFFFF}\textbf{16.43 (3.17)} &
  \cellcolor[HTML]{FFFFFF}\textbf{.058 (.013)} &
   \\ \bottomrule
\end{tabular}
\end{table}

\begin{figure}[H] 
    \centering
    \includegraphics[width=\linewidth, height=7cm]{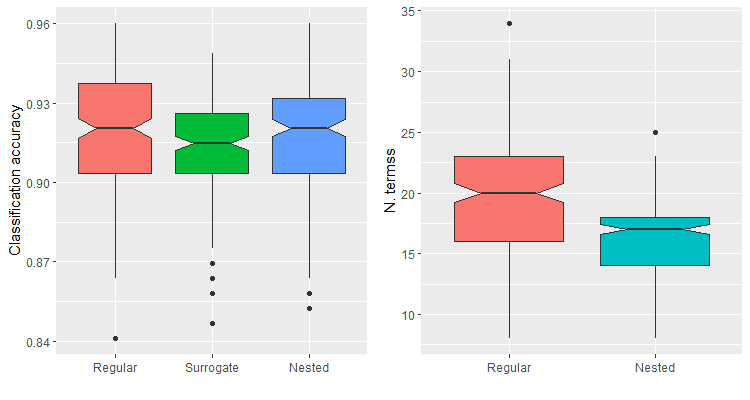}
        \caption{{Boxplots of the distributions of accuracy CCR, and number of terms, per Lasso type. Notches represent the interval covered by$1.58\times\frac{IQR}{\sqrt N} $, with IQR denoting interquartile range.}}
\end{figure}

\subsection{Experiment 7: Metal vs Rock (Classification)}
All 200 iterations worked without issues. Means and standard deviations of accuracy CCR, number of terms, and average accuracy CCR increment per term are depicted in Table 8. The regular Lasso had the highest test accuracy, while the surrogate and nested Lassos did not significantly differ from each other. For example, the 95\% confidence interval of the differences between the regular and nested Lassos accuracy CCR ranged from .006 to .016. The regular Lasso also had a greater number of terms than the nested Lasso (95\% CI [8.33-9.90]), and thus the nested Lasso had higher CCR increment per term (95\% CI [.0104-.0128]). The distributions of accuracy CCR and number of terms are shown, per method, in Figure 7.

\begin{table}[H]
\centering
\caption{{Summary of results based on the Metal vs Rock dataset. Values represent means over 200 iterations (standard deviations in parentheses). In each column, the result of the best performing method (excluding boosting) is printed in bold.
}}
\begin{tabular}{@{}lcccl@{}}
\toprule
\cellcolor[HTML]{FFFFFF}{\color[HTML]{333333} \textbf{Procedure}} &
  \cellcolor[HTML]{FFFFFF}\textbf{Accuracy CCR} &
  \cellcolor[HTML]{FFFFFF}\textbf{Terms} &
  \cellcolor[HTML]{FFFFFF}\textbf{Accuracy CCR per term} &
   \\ \midrule
\textbf{Boosting}        & .779  (.038)         & 1468 (189)            & NA           &  \\
\textbf{Regular Lasso}   & \textbf{.774 (.042)} & {29.32 (5.86)} & .027 (.0057) &  \\
\textbf{Surrogate Lasso} & {.762 (.035)} & 100.41 (4.97)         & .008 (.0005) &  \\
\cellcolor[HTML]{FFFFFF}\textbf{Nested Lasso} &
  \cellcolor[HTML]{FFFFFF}{.763 (.044)} &
  \cellcolor[HTML]{FFFFFF}\textbf{20.17 (3.67)} &
  \cellcolor[HTML]{FFFFFF}\textbf{.039 (.0076)} &
   \\ \bottomrule
\end{tabular}
\end{table}

\begin{figure}[H] 
    \centering
    \includegraphics[width=\linewidth, height=7cm]{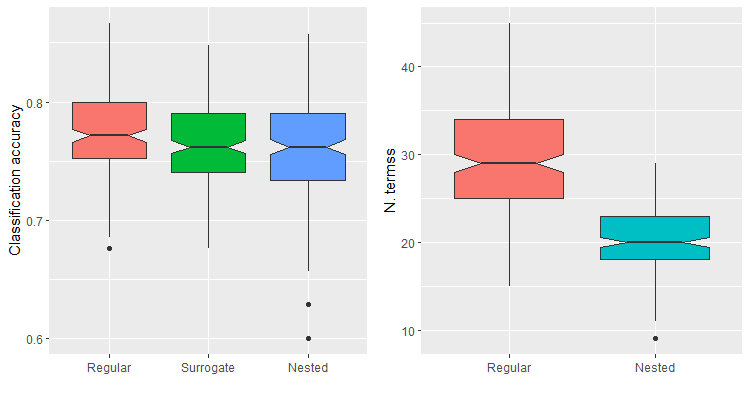}
        \caption{{Boxplots of the distributions of accuracy CCR, and number of terms, per Lasso type. Notches represent the interval covered by$1.58\times\frac{IQR}{\sqrt N} $, with IQR denoting interquartile range.}}
\end{figure}

\section{Discussion}
This study set out to assess whether model-based data-generation would improve the performance of PRE. Neither the surrogate nor the nested Lassos consistently resulted in improved, nor reduced accuracy compared to the regular Lasso. Yet the nested Lasso reliably produced simpler models, resulting in improved quality of terms.

The surrogate Lasso often resulted in accuracy improvements in the regression problems, compared with the other Lasso approaches. With the exception of experiment 4, in which the boosting model, which was mimicked by the surrogate Lasso, was less accurate than PRE with regular/nested Lasso. Unexpectedly, the surrogate Lasso did slightly worse than the other Lassos across the classification problems, regardless of the boosting model's relative accuracy. One might speculate that this was because the surrogate PRE was the only approach in which the Lasso step solely relied on the logit values predicted by the Boosting model. However, attempts to use class predictions instead of logit values did not lead to improvements in accuracy. In addition the surrogate Lasso consistently yielded much more terms than the other Lassos, leading to lower overall quality of terms across all the prediction tasks. Given these indicators, the surrogate Lasso can only be recommended in regression cases, if the Boosting model performs well, and computational efficiency and model complexity are not of major concern.

The nested Lasso resulted in considerable and reliable improvements in model simplicity and the quality of terms, without consistent costs to accuracy compared to the regular Lasso. Furthermore, we found no evidence that the nested approach harmed the validity of variable importance measures, as seen in the high correlations observed between variable importances and assigned slopes in experiment 1. However, it should be noted that the nested Lassos' increased algorithmic dependence on the Boosting model implies that the nested Lasso might be harmed by relatively poor performance by the Boosting model, as seen in experiment 4. Given these results, the nested Lasso is recommended in cases where model simplicity is valued, PRE's Boosting step performs well, and the increased computational demands are tolerable.

\subsection{Limitations and future work}
Algorithmic stability and consistency were hypothesized to improve under the surrogacy-based Lasso approaches, but were not directly assessed, and thus their effects could not be quantified. There is a need to further examine the mechanism for, and the conditions under which improved accuracy can be expected from the surrogate Lasso, and improved simplicity can be expected from the nested Lasso. Further understanding of these mechanisms would be particularly useful given the potential to use model-based data-generation alongside alternative Lasso algorithms, which were designed to improve the Lasso's stability, or its consistency. For example, \cite{LimyYu16} proposed a more stable Lasso algorithm for regression problems, which, similarly to the nested Lasso, resulted in greater simplicity and better terms, at no cost to accuracy. Furthermore, \cite{LiyShao15} alternative Lasso algorithm for regression problems was shown to be significantly more consistent than the standard Lasso when $N < P$, which resulted in higher sensitivity in variable selection. Thus, future work could be aimed at comparing alternative Lasso algorithms to the surrogate and nested Lassos, and even at combining the algorithms in an attempt to improve the performance of PRE.

It should be noted that the results reported here are unlikely to represent the optimal outcomes achievable by the different approaches. Modeling parameters such as the Boosting model's learning rate, the size of the generated datasets, and the Lasso's $\lambda$ could have been more fine-tuned for each case, which would likely have led to improved performance. Furthermore, data-generation methods producing fewer unjustifiable counterfactuals - synthetic data that are unlikely to occur - could have done better work than the simple data-generation algorithm we employed \cite{LaugyLeso19}. However, the purpose of this article was to be a starting point for the examination of model-based data generation in the context of PRE, and to compare the different approaches under similar conditions, rather than maximize performance in each case. 

Lastly, the model-based data generation procedures in this paper only relied on Boosting models. Although the pedagogical surrogacy approach taken here should be model-agnostic \cite{GuidyMonr18}, some models might still be more difficult to mimic than others. However, the strength of the pedagogical approach is that, in theory, every model can be mimicked given a large enough generated dataset. Thus, while it would be beneficial to assess the different Lasso approaches given a greater variety of data-generators, this goes beyond the scope of this paper, which aimed to provide a first assessment of the usefulness of using simple model-based data generation within PRE.

\subsection{Conclusion} This study showed that model-based data generation can indeed improve the performance of PRE. The surrogate Lasso can provide improved regression accuracy, at a cost of increased complexity and computational demands. Whereas the nested Lasso should lead to lower complexity, at minimal costs to accuracy, thereby easing the accuracy-interpretability tradeoff at the cost of greater computational demands. Given these findings, model-based data generation seems to be a promising approach to improve PRE's interpretability, or in some cases its accuracy, which would benefit from further study and refinement.

\newpage

\bibliographystyle{plain} 
\bibliography{template} 

\begin{thebibliography}{10}

\bibitem{AliyTibs19}
Alnur Ali and Ryan~J Tibshirani.
\newblock The generalized lasso problem and uniqueness.
\newblock {\em Electronic Journal of Statistics}, 13(2):2307--2347, 2019.

\bibitem{BuhlyHoth07}
Peter B{\"u}hlmann and Torsten Hothorn.
\newblock Boosting algorithms: Regularization, prediction and model fitting.
\newblock {\em Statistical Science}, 22(4):477--505, 2007.

\bibitem{CravyShav95}
Mark Craven and Jude Shavlik.
\newblock Extracting tree-structured representations of trained networks.
\newblock {\em Advances in Neural Information Processing Systems}, 8:24--30,
  1995.

\bibitem{DeBiyJani16}
Riccardo De~Bin, Silke Janitza, Willi Sauerbrei, and Anne-Laure Boulesteix.
\newblock Subsampling versus bootstrapping in resampling-based model selection
  for multivariable regression.
\newblock {\em Biometrics}, 72(1):272--280, 2016.

\bibitem{DormyElit13}
Carsten~F Dormann, Jane Elith, Sven Bacher, Carsten Buchmann, Gudrun Carl,
  Gabriel Carr{\'e}, Jaime R~Garc{\'\i}a Marqu{\'e}z, Bernd Gruber, Bruno
  Lafourcade, Pedro~J Leit{\~a}o, et~al.
\newblock Collinearity: a review of methods to deal with it and a simulation
  study evaluating their performance.
\newblock {\em Ecography}, 36(1):27--46, 2013.

\bibitem{Fokk20}
Marjolein Fokkema.
\newblock Fitting prediction rule ensembles with {R} package pre.
\newblock {\em Journal of Statistical Software}, 92(1):1--30, 2020.

\bibitem{Frie01}
Jerome~H Friedman.
\newblock Greedy function approximation: A gradient boosting machine.
\newblock {\em Annals of Statistics}, pages 1189--1232, 2001.

\bibitem{Frie02}
Jerome~H Friedman.
\newblock Stochastic gradient boosting.
\newblock {\em Computational Statistics \& Data Analysis}, 38(4):367--378,
  2002.

\bibitem{FrieyPope08}
Jerome~H Friedman, Bogdan~E Popescu, et~al.
\newblock Predictive learning via rule ensembles.
\newblock {\em Annals of Applied Statistics}, 2(3):916--954, 2008.

\bibitem{Fu98}
Wenjiang~J Fu.
\newblock Penalized regressions: the bridge versus the lasso.
\newblock {\em Journal of Computational and Graphical Statistics},
  7(3):397--416, 1998.

\bibitem{GuidyMonr18}
Riccardo Guidotti, Anna Monreale, Salvatore Ruggieri, Franco Turini, Fosca
  Giannotti, and Dino Pedreschi.
\newblock A survey of methods for explaining black box models.
\newblock {\em ACM Computing Surveys}, 51(5):1--42, 2018.

\bibitem{HothyHorn06}
Torsten Hothorn, Kurt Hornik, and Achim Zeileis.
\newblock Unbiased recursive partitioning: A conditional inference framework.
\newblock {\em Journal of Computational and Graphical Statistics},
  15(3):651--674, 2006.

\bibitem{HothyZeil15}
Torsten Hothorn and Achim Zeileis.
\newblock partykit: A modular toolkit for recursive partytioning in {R}.
\newblock {\em The Journal of Machine Learning Research}, 16(1):3905--3909,
  2015.

\bibitem{LaugyLeso19}
Thibault Laugel, Marie-Jeanne Lesot, Christophe Marsala, Xavier Renard, and
  Marcin Detyniecki.
\newblock Unjustified classification regions and counterfactual explanations in
  machine learning.
\newblock In {\em Joint European Conference on Machine Learning and Knowledge
  Discovery in Databases}, pages 37--54. Springer, 2019.

\bibitem{LeisyDimi21}
Friedrich Leisch and Evgenia Dimitriadou.
\newblock {\em mlbench: Machine Learning Benchmark Problems}, 2021.
\newblock R package. https://CRAN.R-project.org/package=mlbench.

\bibitem{LiyShao15}
Quefeng Li and Jun Shao.
\newblock Regularizing lasso: a consistent variable selection method.
\newblock {\em Statistica Sinica}, 25(3):975--992, 2015.

\bibitem{LimyYu16}
Chinghway Lim and Bin Yu.
\newblock Estimation stability with cross-validation {(ESCV)}.
\newblock {\em Journal of Computational and Graphical Statistics},
  25(2):464--492, 2016.

\bibitem{Mein10}
Nicolai Meinshausen.
\newblock Node harvest.
\newblock {\em The Annals of Applied Statistics}, 4(4):2049--2072, 2010.

\bibitem{MingQu18}
Yao Ming, Huamin Qu, and Enrico Bertini.
\newblock Rulematrix: Visualizing and understanding classifiers with rules.
\newblock {\em IEEE Transactions on Visualization and Computer Graphics},
  25(1):342--352, 2018.

\bibitem{NogeySechi17}
Sarah Nogueira, Konstantinos Sechidis, and Gavin Brown.
\newblock On the stability of feature selection algorithms.
\newblock {\em Journal of Machine Learning Research}, 18(1):6345--6398, 2017.

\bibitem{Quin14}
J~Ross Quinlan.
\newblock {\em C4.5: Programs for machine learning}.
\newblock Morgan Kaufman Publishers, San Matea, CA, 2014.

\bibitem{ShimyLi14}
Toshio Shimokawa, Li~Li, Kun Yan, Shinnichi Kitamura, and Masashi Goto.
\newblock Modified rule ensemble method for binary data and its applications.
\newblock {\em Behaviormetrika}, 41(2):225--244, 2014.

\bibitem{Sigr21}
Fabio Sigrist.
\newblock Gradient and newton boosting for classification and regression.
\newblock {\em Expert Systems With Applications}, 167:114080, 2021.

\bibitem{Tibs96}
Robert Tibshirani.
\newblock Regression shrinkage and selection via the lasso.
\newblock {\em Journal of the Royal Statistical Society: Series B
  (Methodological)}, 58(1):267--288, 1996.

\bibitem{XuyCara11}
Huan Xu, Constantine Caramanis, and Shie Mannor.
\newblock Sparse algorithms are not stable: A no-free-lunch theorem.
\newblock {\em IEEE Transactions on Pattern Analysis and Machine Intelligence},
  34(1):187--193, 2011.

\bibitem{YarkyWest17}
Tal Yarkoni and Jacob Westfall.
\newblock Choosing prediction over explanation in psychology: Lessons from
  machine learning.
\newblock {\em Perspectives on Psychological Science}, 12(6):1100--1122, 2017.

\end{thebibliography}

\newpage
\section*{Appendix}

\begin{table}[H]
\caption{{Average computation time, and variable selection stability of PREs. In each row, the result of the best performing method is printed in bold.} }
\begin{tabular}{@{}
>{\columncolor[HTML]{FFFFFF}}l cccl@{}}
\toprule
{\color[HTML]{333333} \textbf{Source}} &
  \cellcolor[HTML]{FFFFFF}\textbf{Regular PRE} &
  \cellcolor[HTML]{FFFFFF}\textbf{PRE with surrogate Lasso} &
  \cellcolor[HTML]{FFFFFF}\textbf{PRE with nested Lasso} &
   \\ \midrule
\textbf{Computation time}             & \cellcolor[HTML]{FFFFFF}               & \cellcolor[HTML]{FFFFFF}              & \cellcolor[HTML]{FFFFFF}      &  \\
{Exp. 1, Simulation} & \textbf{7.44}                          & {12.90}                        & 11.96                         &  \\
{Exp. 2, Boston Housing}       & \textbf{19.52}                         & 40.38                                 & 34.82                         &  \\
{Exp. 3, Metal vs. Rock} &
  \cellcolor[HTML]{FFFFFF}\textbf{43.09} &
  \cellcolor[HTML]{FFFFFF}{57.11} &
  \cellcolor[HTML]{FFFFFF}{53.00} &
     \\
Exp. 4, Breast Cancer*          & \textbf{18.57}                         & 33.87                                 & 29.66                         &  \\
Exp. 5, Ionosphere*             & \textbf{35.37}                         & 52.67                                 & 47.73                         &  \\
Exp. 6, Metal vs. Rock*         & \cellcolor[HTML]{FFFFFF}\textbf{29.36} & \cellcolor[HTML]{FFFFFF}38.5          & \cellcolor[HTML]{FFFFFF}35.89
   \\ \midrule
\textbf{Variable selection stability} & \cellcolor[HTML]{FFFFFF}               & \cellcolor[HTML]{FFFFFF}              & \cellcolor[HTML]{FFFFFF}      &  \\
Exp. 1, Simulation                    & NA                                     & NA                                    & NA                            &  \\
Exp. 2, Boston Housing                & .75                                    & \textbf{.84}                          & .71                           &  \\
Exp. 3, Metal vs. Rock                & NA                                     & NA                                    & NA                            &  \\
Exp. 4, Breast Cancer*                & .132                                   & -.001                                 & \textbf{.138}                 &  \\
Exp. 5, Ionosphere*                   & .46                                    & \textbf{.56}                          & .52                           &  \\
Exp. 6, Metal vs. Rock*               & \cellcolor[HTML]{FFFFFF}.252           & \cellcolor[HTML]{FFFFFF}\textbf{.362} & \cellcolor[HTML]{FFFFFF}.356  & 
\\ \bottomrule
{ Note. Asterisks (*) indicate classification problems, while the lack of asterisk indicates regression problems. }\\
\end{tabular}
\end{table}

\end{document}